\documentclass[10pt,twocolumn,letterpaper]{article}

\usepackage{cvpr}              

\usepackage{xcolor}

\definecolor{cvprblue}{rgb}{0.21,0.49,0.74}
\usepackage[pagebackref,breaklinks,colorlinks,allcolors=cvprblue]{hyperref}

\usepackage{multirow}
\usepackage{booktabs}
\usepackage{adjustbox}
\usepackage{adjustbox}
\usepackage{pifont}  

\def\paperID{} 
\def\confName{CVPR}
\def\confYear{2026}

\title{WhisperNet: A Scalable Solution for Bandwidth-Efficient Collaboration}

\author{
Gong Chen$^{1}$ \quad
Chaokun Zhang$^{2}$\thanks{Corresponding author.} \quad
Xinyan Zhao$^{2}$ 
\\
$^{1}$School of Computer Science and Technology, Tianjin University  \quad \\
$^{2}$School of Cybersecurity, Tianjin University  \quad
}

\begin{document}

\maketitle

\begin{abstract}

Collaborative perception is vital for autonomous driving yet remains constrained by tight communication budgets. Earlier work reduced bandwidth by compressing full feature maps with fixed-rate encoders, which adapts poorly to a changing environment, and it further evolved into spatial selection methods that improve efficiency by focusing on salient regions, but this object-centric approach often sacrifices global context, weakening holistic scene understanding. To overcome these limitations, we introduce \textit{WhisperNet}, a bandwidth-aware framework that proposes a novel, receiver-centric paradigm for global coordination across agents. Senders generate lightweight saliency metadata, while the receiver formulates a global request plan that dynamically budgets feature contributions across agents and features, retrieving only the most informative features. A collaborative feature routing module then aligns related messages before fusion to ensure structural consistency.
Extensive experiments show that WhisperNet achieves state-of-the-art performance, improving AP@0.7 on OPV2V by 2.4\% with only 0.5\% of the communication cost. As a plug-and-play component, it boosts strong baselines with merely 5\% of full bandwidth while maintaining robustness under localization noise. These results demonstrate that 
globally-coordinated allocation across \textit{what} and \textit{where} to share is the key to achieving efficient collaborative perception.


\end{abstract}

\section{Introduction}

Collaborative perception (CP) enhances autonomous driving safety through vehicle-to-everything (V2X) communication \cite{cp} to overcome single-agent limitations like occlusions \cite{li2024bevformer, yang2023bevformer}. 
However, these benefits incur substantial communication overhead, as transmitting raw sensor data or high-dimensional features \cite{chen2019cooper} in real-time consumes a vast amount of bandwidth. This bandwidth constraint has become a critical bottleneck, hindering the large-scale deployment of CP \cite{mao20233d, han2023collaborative}. Therefore, designing communication-efficient methods that maintain high perception accuracy under strict bandwidth budgets remains a challenge.



To mitigate bandwidth, prior work mainly follows two strategies. First, \textbf{feature compression \& reconstruction methods} \cite{wang2023core, codefilling} encode the full feature map into a compact latent and decode it at the receiver; however, fixed-rate encoders or static codebooks are inflexible, failing to adapt to dynamic bandwidth and varying scene complexity \cite{zhang2025dsrc}, 
thus enforcing a fidelity-bandwidth compromise. 
To gain more flexibility, \textbf{spatial selection methods} \cite{hu2022where2comm, zhang2024ermvp} are proposed to reduce bandwidth by transmitting only salient spatial features. For instance, methods like Where2comm \cite{hu2022where2comm} and CoRA \cite{chen2025cora} prioritize foreground objects. 

\begin{figure}[tp] 
  \centering 
  \includegraphics[width=1.0\linewidth]{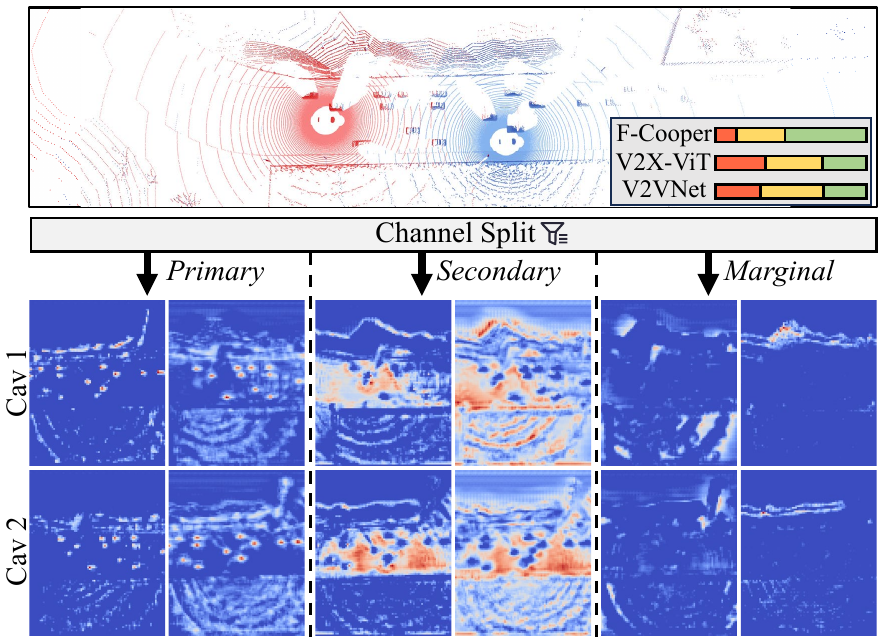} 
  \caption{Visualization of different channel groups. The proportional table in the figure represents the model's primary (Orange), secondary (Yellow), and marginal (Green) channels.}
  \label{fig:abstract} 
\end{figure}

\begin{figure}[tp] 
  \centering 
  \includegraphics[width=1.0\linewidth]{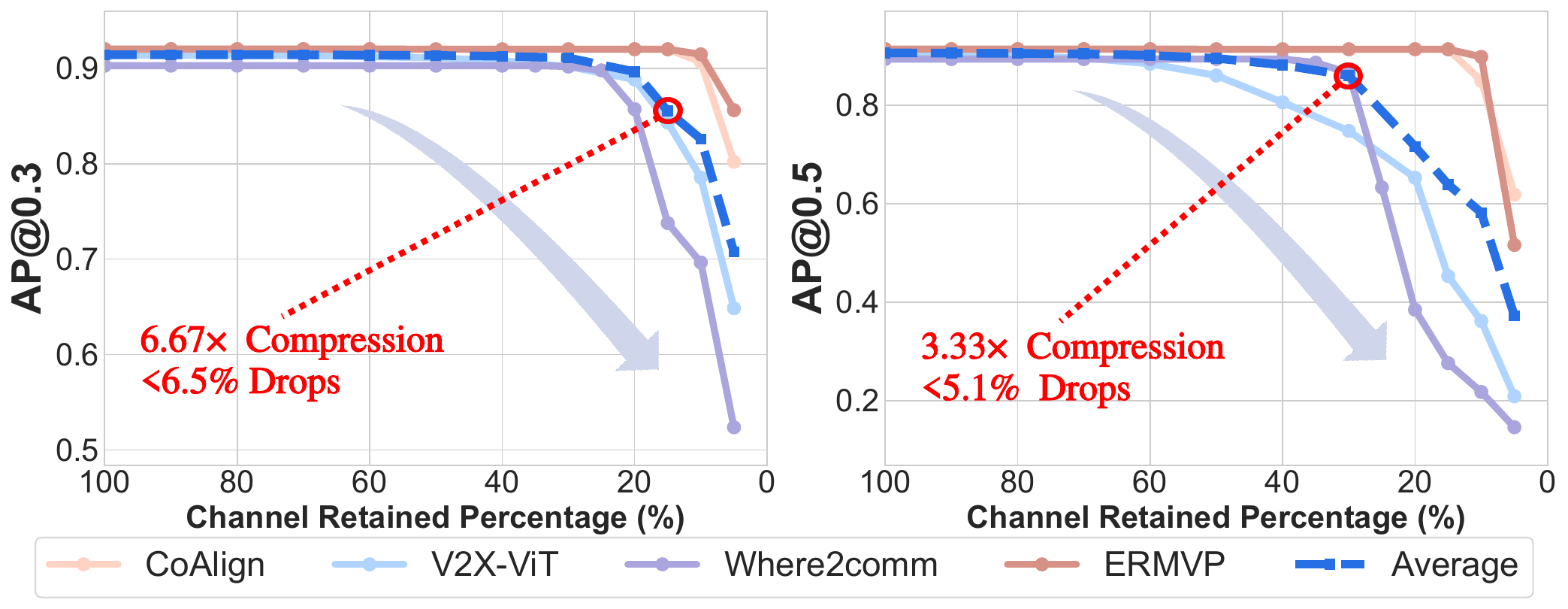}
  \caption{Impact of Channel Pruning on Performance. }
  \label{fig:abstract2} 
\end{figure}

However, while spatial-only methods optimize where to communicate, they still transmit full, redundant channels. Even recent filtering methods \cite{yang2023how2comm} may not fully exploit this redundancy on a global, system-wide scale. 
This reveals a critical gap: existing methods have overlooked the significant redundancy present in the channel dimension.
This gap motivates our investigation into the channel properties. As shown in Fig.~\ref{fig:abstract}, channels can be qualitatively grouped into three categories: primary channels, which are crucial for defining objects; secondary channels, which provide complementary contextual details; marginal channels, which are largely redundant. Transmitting these marginal channels not only wastes valuable bandwidth but can also introduce noise that degrades perception accuracy.
We validate this hypothesis with a preliminary experiment. By ranking channels based on their post-fusion $L1$-norm \cite{wu2018l1} and pruning the unimportant ones, we observe a striking result (Fig.~\ref{fig:abstract2}): Removing nearly half of the channels does not harm performance. 
This finding confirms the unexploited potential of channel redundancy and underscores the need for a more principled, content-aware allocation strategy.

While prior methods have mainly focused on selecting spatial regions, our findings on channel redundancy reveal that this addresses only half of the problem. This leads to our core insight: an effective communication strategy must jointly optimize \textit{where} to communicate and \textit{what} to transmit.
To this end, we present WhisperNet, a novel framework that actualizes this optimization via a globally-coordinated allocation scheme. This scheme formulates a holistic plan for feature exchange based on system-wide metadata, ensuring a complementary fusion. WhisperNet comprises three key modules.
1) Sender-side Importance Estimation Module, which employs a lightweight attention mechanism to jointly estimate spatial and channel importance maps for transmission; 
2) Receiver-side Confidence-Aware Module, which acts as the global coordinator, gathering all metadata to formulate a global request plan that dynamically budgets feature contributions across agents and features;
3) Collaborative Feature Routing Module, which aligns the received messages by directing related subsets to specialized sub-networks, enabling context-aware integration.
Experiments demonstrate WhisperNet's superiority. When limited to only 1\% bandwidth, it surpasses the second-best method's accuracy by 10.6\%/11.7\% on OPV2V/DAIR-V2X.
Moreover, its plug-and-play design enables seamless integration into existing models, reducing their communication cost by 95\% on average with negligible accuracy loss.

Our contributions are summarized as follows: (1) We tackle the critical problem of redundancy in CP by proposing a novel paradigm that shifts from sender-side filtering to receiver-centric global coordination; (2) We design WhisperNet, a lightweight module that can integrate with existing methods to achieve significant communication reduction at a negligible performance cost. (3) Extensive experiments verify our state-of-the-art approach for performance-bandwidth trade-off under strict constraints.

\section{Related Work}

\noindent \textbf{Collaborative Perception.}
CP enables agents to achieve perceptual capabilities beyond those of a single vehicle by sharing messages. Mainstream approaches fall into three categories: Early fusion \cite{MOT_CUP, chen2019cooper} directly transmits raw perception data, offering high accuracy but incurring immense bandwidth overhead. Methods like Cooper \cite{chen2019cooper}, which performs 3D object detection by aggregating multi-vehicle point clouds. Late fusion \cite{Quest, Coop3D} minimizes the communication burden by only exchanging detection results, but its robustness is limited under occlusion and in blind spots. For instance, Coop3D \cite{Coop3D} implements vehicle-infrastructure collaborative detection via a result-level sharing method. Positioned between these two, intermediate fusion \cite{xu2022v2xvit, yang2023how2comm, tang2025rocooper} strikes a balance between accuracy and bandwidth by sharing intermediate features. Among these methods, V2X-ViT \cite{xu2022v2xvit} utilizes heterogeneous multi-agent attention and multi-scale window attention to model cross-agent interaction and spatio-temporal alignment, while V2VNet \cite{wang2020v2vnet} implements feature-level fusion with a graph attention and message-passing mechanism. Nevertheless, the bandwidth required by intermediate fusion schemes remains a key bottleneck, which constrains their large-scale deployment.

\begin{figure*}[ht] 
  \centering 
  \includegraphics[width=1.0\linewidth]{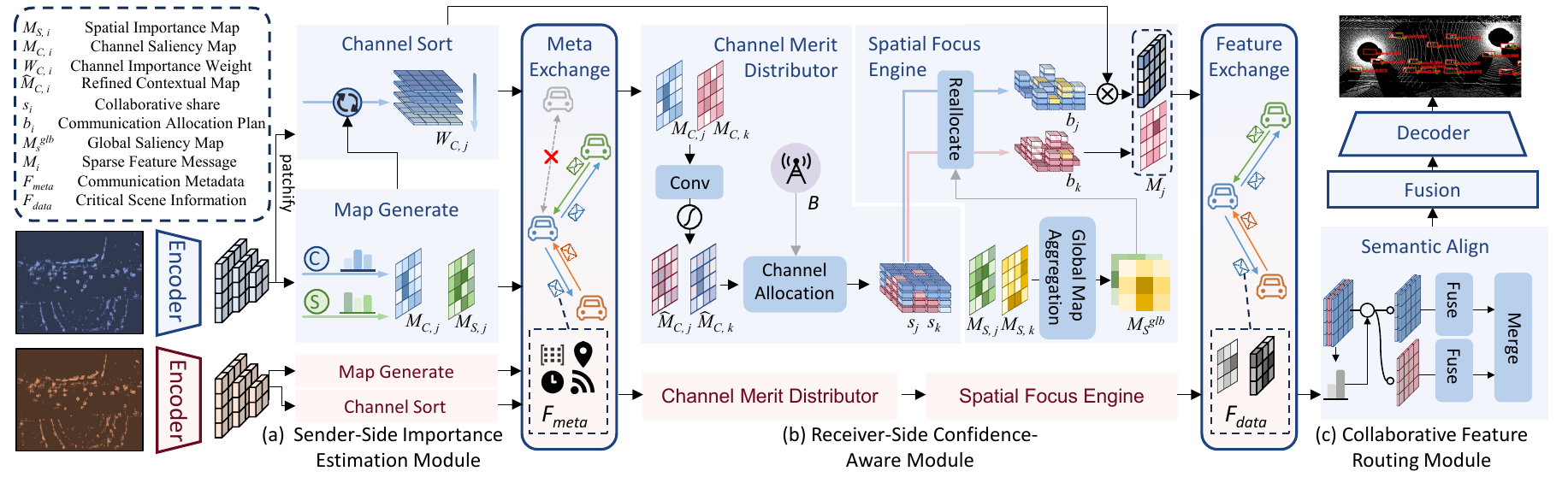}
  \caption{
  The architecture of WhisperNet. Each sender agent first extracts features from its raw sensor data and generates compact importance maps that summarize both spatial and channel-wise saliency. These maps are then exchanged with a receiver agent, which dynamically 
  requests only the most critical feature subsets. Finally, the requested features are fused by a collaborative routing module.
  }
  \label{fig:overall} 
\end{figure*}

\noindent \textbf{Multi-agent Communication.}
Earlier approaches reduce bandwidth by holistically compressing intermediate features on the sender and reconstructing them on the receiver. Typical designs include cooperative reconstruction with a compressor-decoder like AttFuse \cite{xu2022opv2v} and CORE \cite{wang2023core}. While effective at lowering volume, these schemes rely on static encoders or codebooks \cite{codefilling,CoGMP}, limiting adaptability to dynamic bandwidth and scene complexity.
Who2com \cite{liu2020who2com} and When2com \cite{liu2020when2com} line learns which partners to talk to and when to form groups, typically sending full features once selected. This reduces traffic via selective partner/time scheduling but struggles to scale as the number of agents grows.
Recent methods improve the bandwidth-accuracy trade-off via two primary strategies: one-stage \cite{zhang2024ermvp, yang2023how2comm} or two-stage \cite{hu2022where2comm,CoSDH, chen2025cora} protocols. One-stage spatial methods directly send salient tiles. For example, How2comm \cite{yang2023how2comm} combines feature filtering and pragmatic fusion for direct transmission. Two-stage protocols first exchange needs, then transmit features accordingly. For example, Where2comm \cite{hu2022where2comm} introduced spatial confidence maps to send only salient regions; subsequent CoSDH \cite{CoSDH} models supply-demand hybridization. However, these methods remain largely ego-centric or pairwise in their decision-making. For instance, CoSDH's pairwise allocation can still miss the global optimum, leading to duplicated uploads for the same region or, conversely, gaps where no one uploads. This motivates a globally coordinated scheme that jointly decides, for every region, who should transmit and how to avoid redundancy while preserving scene completeness.

\section{Methods}

\subsection{Methodology Overview}

\noindent \textbf{Problem Formulation.} 
We consider a system of $N$ agents, $\mathcal{V} = \{1, \dots, N\}$, including the ego agent $i$, operating over $K$ communication rounds. In each round $k \in \{1, \dots, K\}$, each agent $j \in \mathcal{V}$ extracts a local feature map $X_j$ from its sensors. The agent then utilizes its communication module $\mathcal{C}$ to generate and exchange a set of messages, denoted as $\mathbb{M}_j^{(k)}$ with other agents. The total communication process is constrained by a bandwidth budget $B_j$ for each agent. 
The goal is to design an optimal communication module $\mathcal{C}$ that maximizes the perception performance $\mathcal{P}$ for any agent $j$ with fusion module $\mathcal{F}$ under the given bandwidth constraints. This optimization problem is expressed as:
\begin{equation}
\begin{aligned}
\underset{\mathcal{C}}{\text{maximize}} \quad
& \mathcal{P}\!\left(\mathcal{F}\!\left( X_i, \{ \mathbb{M}_{j}^{(k)} \}_{j \neq i, \forall k} \right)\right) \\[-6pt]
\text{s.t.} \quad
& \sum_{k=1}^{K} \text{Size}(\mathbb{M}_j^{(k)}) \le B_j,\; \forall j \in \mathcal{V}
\end{aligned}
\label{eq:eq1}
\end{equation}

\noindent \textbf{Framework Overview.} 
As shown in Fig.~\ref{fig:overall}, WhisperNet implements the communication module $\mathcal{C}$ through three modules. 
1) Sender-Side Importance Estimation Module: This module analyzes the local feature map to generate a spatial-channel importance map that summarizes the agent's information. 
2) Receiver-Side Confidence-Aware Module: It gathers importance maps from other agents, creating a dynamic communication plan and requesting the most critical data across all agents.
3) Collaborative Feature Routing Module: 
This module performs channel-wise alignment of the sparse features from all agents, serving as a preparation step for downstream fusion.

\subsection{Sender-Side Importance Estimation Module}

The role of the Sender-Side Importance Estimation module is to analyze the local feature $X_j \in \mathbb{R}^{H \times W \times C}$ and generate a joint importance map for transmission that guides the coordination by highlighting the spatial and channel-wise locations of the most salient features. 
The module's operation is divided into two primary functions: Map Generation to produce importance maps for transmission and Channel Sort to prepare local channel weights for selection.

\begin{figure}[t] 
  \centering 
  \includegraphics[width=1.0\linewidth]{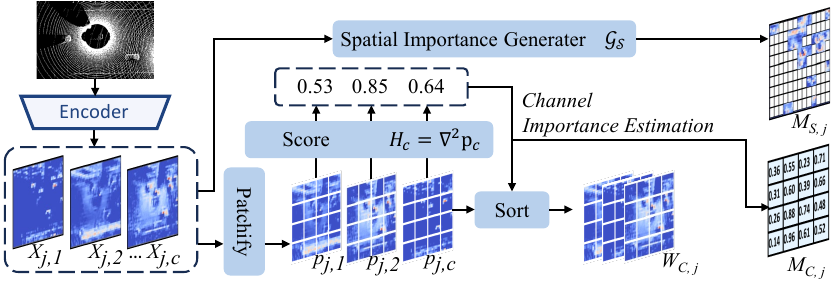}
  \caption{Sender-Side Importance Estimation module. It evaluates feature from both spatial and channel perspectives to generate importance maps that guides communication coordination.}
  \label{fig:m1} 
\end{figure}

\noindent \textbf{Map Generation.}
This function generates two distinct importance maps for transmitting: a spatial importance map $M_{S,j}$ and a channel saliency map $M_{C,j}$.
As shown in Fig.~\ref{fig:m1}, the Spatial Importance Predictor identifies critical regions within the feature. It takes the entire feature $X_j$ as input and outputs a spatial importance $M_{S,j} \in [0,1]^{H \times W}$, where each value indicates the significance of a spatial location. This is achieved using a mapping function $\mathcal{G}_S$.
\begin{equation}
    M_{S,j}=\mathcal{G}_S(X_j)
\end{equation}
where $\mathcal{G}_S$ is implemented as a two-layer convolutional head. 

Channel Importance Estimation determines what content is most informative. Motivated by prior work \cite{high1,high2}, which indicates high-frequency details are more valuable than low-frequency features, our method determines this by scoring the information density of each feature channel.

{
We first score channels by patch-wise information density.
For the $c$-th channel $X_{j,c}\in\mathbb{R}^{H\times W}$, we divide it into
patches $\{p_{j,c}^k\}_{k=1}^{Q}$ with $Q=\frac{H}{P}\cdot\frac{W}{P}$ to preserve feature coherency. We then compute a Laplacian magnitude score for each patch to derive its informativeness:
\begin{equation}
S(p_{j,c}^k)=\|\nabla^2 p_{j,c}^k\|_1
=\sum_{(u,v)\in \mathcal{P}_k}|(\mathcal{L}\ast X_{j,c})_{u,v}|
\end{equation}
where $\mathcal{L}$ is a 3 $\times$ 3 Laplacian kernel and 
$\mathcal{P}_k$ denotes the pixel index set of the $k$-th patch.
From these patch scores, we then generate a 2D channel saliency map $M_{C,j}$. 
}


For each patch location $k$, we classify channels into three groups primary, secondary, and marginal using a lightweight $1{\times}1$ conv head to generate three classification scores for each channel, followed by a softmax over the three groups, yielding the resulting group probabilities $\boldsymbol{\pi}_{j,k,c}=[\pi^{\text{pri}}_{j,k,c},\,\pi^{\text{sec}}_{j,k,c},\,\pi^{\text{mar}}_{j,k,c}]^{\top}$ for channel $c$. We further let 
$\boldsymbol{\omega}=[\omega_{\text{pri}},\,\omega_{\text{sec}},\,\omega_{\text{mar}}]^{\top}$
be learnable group weights. 
We compute the group-weighted channel saliency map $M_{C,j}$ for transmission by taking the normalized maximum score at each patch $k$:
\begin{equation}
{M}_{C,j}(k)=
\max_c \!\left[
(\boldsymbol{\omega}^{\!\top}\!\boldsymbol{\pi}_{j,k,c})\,S(p^k_{j,c})
\right].
\end{equation}

\noindent \textbf{Channel Sort.} In parallel, we form a channel weights for local use. For patch $k$, we collect channel scores into a vector 
$\mathbf{s}_{j,k} $ $= [S(p_{j,1}^k),\dots,S(p_{j,C}^k)]^\top$ and compute weights.
\begin{equation}
\mathbf{w}_{j,k}=\mathrm{Softmax}(\mathbf{s}_{j,k})
\end{equation}

The collection of these weights $W_{C,j}=\{\mathbf{w}_{j,k}\}_{k=1}^Q$ is retained by the sender and not transmitted. These weights will be used later by the coordinator to select the most salient channels.
Finally, the importance maps $M_{S,j}$ and $M_{C,j}$ are transmitted to the receivers for further processing.

\subsection{Receiver-Side Confidence-Aware Module}


As depicted in the middle of the Fig. \ref{fig:overall}, the Receiver-Side Confidence-Aware Module works as the ego-centric coordinator and is responsible for intelligently distributing the communication budget $B$ in each cycle.
It adopts a global request-response protocol. The process begins when the ego receives the spatial maps $\{M_{S,j}\}_{j \neq i}$ and channel maps $\{M_{C,j}\}_{j \neq i}$ from all collaborators.

To assess the relative importance of each collaborator's information at different locations, the coordinator employs a Channel Merit Distributor. The distributor refines the raw channel saliency maps using a function $\mathcal{H}$:
\begin{equation}
    \hat{M}_{C,j} = \mathcal{H}(\{M_{C,l}\}_{l \neq i})
\end{equation}
Here, $\mathcal{H}$ consists of a convolutional layer followed by a Gaussian filter, enabling inter-agent interaction and smoothing. The refined map $\hat{M}_{C,j}$ represents the contextual importance of agent $j$. Based on this, the collaborative share $s_{j,k}$ for vehicle $j$ at patch $k$ is defined as:
\begin{equation}
    s_{j,k} = \frac{\hat{M}_{C,j}(k)}{\sum_{l \neq i} \hat{M}_{C,l}(k) + \epsilon}
\end{equation}

However, allocating the budget uniformly across all patches and then distributing it by shares $s_{j,k}$ is suboptimal, as background regions require fewer resources than critical areas. To address this, we introduce a Spatial Focus Engine, which redistributes the budget according to spatial importance maps $\{M_{S,j}\}_{j \neq i}$.
We first generate a unified global saliency map $M_S^{\text{global}}$ by taking the element-wise maximum of all individual spatial maps. This global map is then transformed into a spatial budget distribution $P_S$:
\begin{equation}
    P_S(k) = \frac{\exp(M_S^{\text{global}}(k)/\tau_s)}{\sum_{k'=1}^Q \exp(M_S^{\text{global}}(k')/\tau_s)}
\end{equation}

With $P_S(k)$, the budget for each patch $k$ is computed as:
\begin{equation}
    \hat{B}_{\text{patch}}(k) = B \cdot P_S(k)
\end{equation}



For each patch $k$, the coordinator first allocates the patch budget $\hat{B}_{\text{patch}}(k)$ to all agents in proportion to their collaborative shares $s_{j,k}$:
\begin{equation}
b_{j,k}=\left\lfloor \hat{B}_{\text{patch}}(k)\cdot s_{j,k}\right\rceil.
\end{equation}

The resulting allocation matrix $\{b_{j,k}\}$ is then broadcast to all collaborators.
Upon receiving the allocation plan, each agent $j$ locally prioritizes its transmission according to the channel groups and $W_{C,j}$: primary channels are sent first, followed by secondary, while marginal channels are only transmitted when remaining bandwidth allows.

\subsection{Collaborative Feature Routing Module}

This module serves as a channel alignment and organization layer that harmonizes the sparse features received from all agents before the subsequent fusion stage. Instead of directly performing feature fusion, the module focuses on routing semantically correlated channels into coherent groups, ensuring that information distributed across different agents is well-aligned for downstream integration.

\begin{figure}[t] 
  \centering 
  \includegraphics[width=1.0\linewidth]{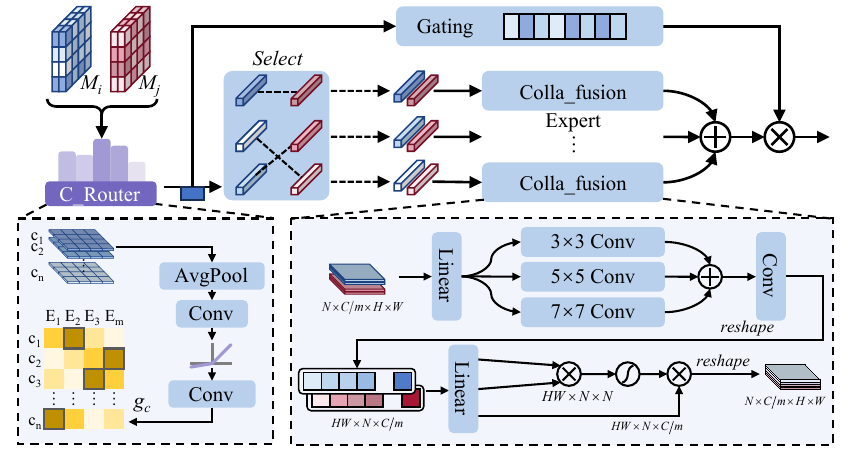}
  \caption{Collaborative Feature Routing module. It enables cross-vehicle feature exchange by grouping and processing semantically similar channels.}
  \label{fig:m3} 
\end{figure}

As shown in Fig.~\ref{fig:m3}, the sparse feature messages $\{M_j\}_{j \in N}$ are first combined into a unified tensor $X_\text{in} \in \mathbb{R}^{N \times C \times H \times W}$ by zero-padding the unselected channels. For each channel $c$, a routing network generates a soft assignment vector $\mathbf{g}_c \in \mathbb{R}^{m}$, indicating its affinity to $m$ specialized experts.
\begin{equation} \mathbf{g}_c = \text{Softmax}(\text{MLP}(\text{GlobalAvgPool}(X_{\text{in}}^c))). \end{equation}

Each input channel is assigned to the expert with the highest affinity, which enables semantic grouping, avoiding indiscriminate direct fusion.

Within each expert, a lightweight multi-scale convolutional block with parallel $3\times3$, $5\times5$, and $7\times7$ kernels refines local structures to normalize and align intra-agent representations, preparing them for consistent downstream processing. An attention-based inter-agent recalibration step is then applied along the agent dimension, adapting per-channel alignment to reduce cross-agent inconsistency. Finally, the outputs of all $m$ experts are recombined through element-wise summation to restore the aligned feature tensor.
This pathway operates in parallel with a residual Squeeze-and-Excitation \cite{se} branch applied to $X_\text{in}$ for global channel recalibration. The output provides a channel-harmonized representation that facilitates stable and efficient fusion in the following Fusion module.

\section{Experiments}


\subsection{Experimental Settings}

\noindent \textbf{Datasets.} We evaluate our method on two large-scale datasets: OPV2V \cite{xu2022opv2v} and DAIR-V2X \cite{dairv2x}.
OPV2V is the first large-scale open dataset for vehicle-to-vehicle perception. The dataset encompasses 73 diverse scenarios across urban and highway environments that simulate realistic traffic and occlusion conditions.
DAIR-V2X is a pioneering large-scale dataset for vehicle-to-infrastructure perception. DAIR-V2X provides synchronized data from both vehicle and infrastructure sensors in real-world traffic scenarios.

\noindent \textbf{Implementation details.}
We follow the standard settings from the OpenCOOD benchmark \cite{xu2022opv2v} for both datasets, with a communication range of 70 meters for all agents. For the 3D object detection task, we restrict the point cloud range to $[-140.8, 140.8] \times [-40, 40] \times [-3, 1]$ meters. We evaluate performance using Average Precision (AP) at Intersection over Union (IoU) thresholds of 0.5 and 0.7 (AP@0.5/AP@0.7). For the semantic segmentation task, the perception range is $[-51.2, 51.2] \times [-51.2, 51.2] \times [-3, 1]$ meters. We report the mean Intersection over Union (mIoU) across the Road, Lane, and Vehicle classes. Across all experiments, we use PointPillars \cite{lang2019pointpillars} as the backbone for feature extraction from point clouds, with a voxel size of 0.4 meters. 
We adopt the default value of 1 from the softmax function for $\tau_s$. The number of experts is set to 4.

Following the OPV2V benchmark, we take the standard intermediate-fusion with $16\times$ compression as the reference, and define it as the communication rate of $1.0$. All reported rates are measured relative to this baseline.

\noindent \textbf{Baselines.}  
We adopt the collaboration model in OPV2V \cite{xu2022opv2v} as the baseline for WhisperNet. In our experiments, WhisperNet is extensively compared with a wide range of state-of-the-art communication-efficient methods, such as Where2comm \cite{hu2022where2comm}, ERMVP \cite{zhang2024ermvp}, and CoSDH \cite{CoSDH}.  
To further assess its plug-and-play capability, WhisperNet is integrated into several representative frameworks: F-Cooper~\cite{fcooper}, V2VNet~\cite{wang2020v2vnet}, and V2X-ViT~\cite{xu2022v2xvit}.  
We evaluate both the communication efficiency and perception performance under varying bandwidth budgets.

\begin{table}[t]
  \centering
  \caption{Performance comparison of mainstream methods.}
  \begin{adjustbox}{max width=\linewidth}
    \begin{tabular}{c|c|cc|cc}
    \hline
    \multirow{2}[2]{*}{\raisebox{1ex}{Methods}} & \multirow{2}[2]{*}{\raisebox{1ex}{Params}} & \multicolumn{2}{c|}{OPV2V} & \multicolumn{2}{c}{DAIR-V2X} \\
    \cline{3-6}
          & & AP@0.5 & AP@0.7 & AP@0.5 & AP@0.7 \\
    \hline
    Single & 6.28 M & 0.8078  & 0.6853  & 0.6250  & 0.4457  \\
    Late  & 6.28 M & 0.8670  & 0.8025  & 0.6971  & 0.5107  \\
    Inter. & 6.28 M & 0.8403  & 0.6774  & 0.6506  & 0.5538  \\
    \hline
    V2VNet \cite{wang2020v2vnet} & 16.08 M & 0.9175  & 0.8221  & 0.6644  & 0.4037  \\
    V2X-ViT \cite{xu2022v2xvit} & 14.50 M & 0.9035  & 0.8119  & 0.7046  & 0.5240  \\
    Where2comm \cite{hu2022where2comm} & 10.43 M & 0.8937  & 0.7889  & 0.6735  & 0.5317  \\
    CoAlign \cite{coalign} & 11.42 M & 0.9132  & 0.8381  & 0.7772  & 0.6284  \\
    How2Comm \cite{yang2023how2comm} & 35.79 M & 0.8589  & 0.7147  & 0.6236  & 0.4718  \\
    MRCNet \cite{mrcnet} & 18.79 M & 0.8529  & 0.7698  & 0.6647  & 0.5385  \\
    DSRC \cite{zhang2025dsrc}  & 10.14 M & 0.9183  & 0.8526  & 0.7852  & 0.6360  \\
    CoSDH \cite{CoSDH} & 8.52 M & 0.8952  & 0.8373  & 0.7042  & 0.5766  \\
    ERMVP \cite{zhang2024ermvp} & 11.87 M & 0.9139  & 0.8404  & 0.7675  & 0.6350  \\
    MDD \cite{mdd}  & 8.09 M & 0.8007  & 0.6817  & 0.7495  & 0.5817  \\
    \hline
    WhisperNet & 7.28 M & 0.9334  & 0.8764  & 0.7915  & 0.6480  \\
    \hline
    \end{tabular}%
  \label{tab:benchmark_comparison}%
  \end{adjustbox}
\end{table}%

\begin{figure*}[ht] 
  \centering 
  \includegraphics[width=1.0\linewidth]{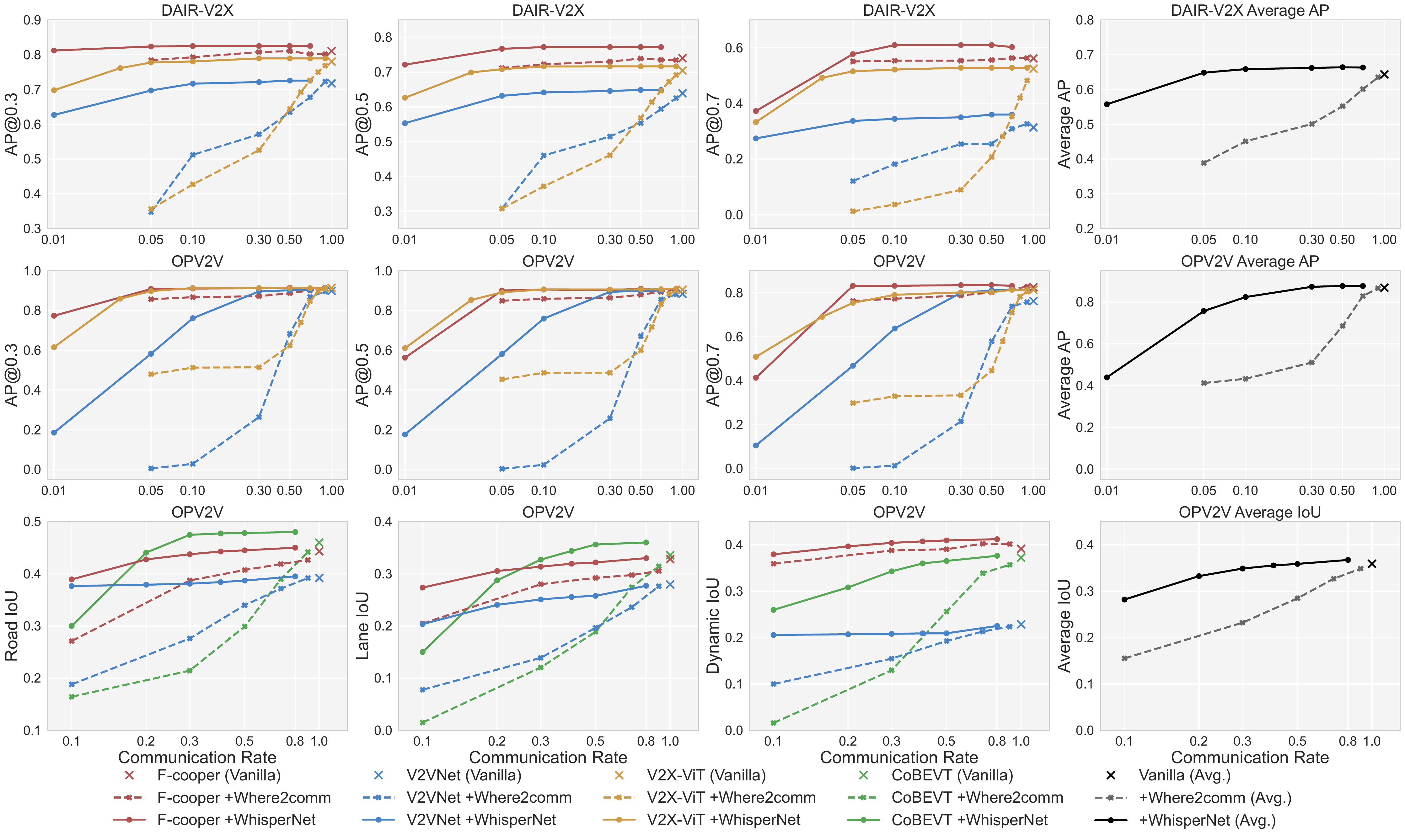} 
  \caption{Plug-and-play performance validation. The figure compares the 3D object detection and BEV semantic segmentation performance when using our method (WhisperNet) and Where2comm as communication plug-ins under varying communication rates.}
  \label{fig:performance} 
\end{figure*}

\begin{figure}[!t] 
  \centering 
  \includegraphics[width=1.0\linewidth]{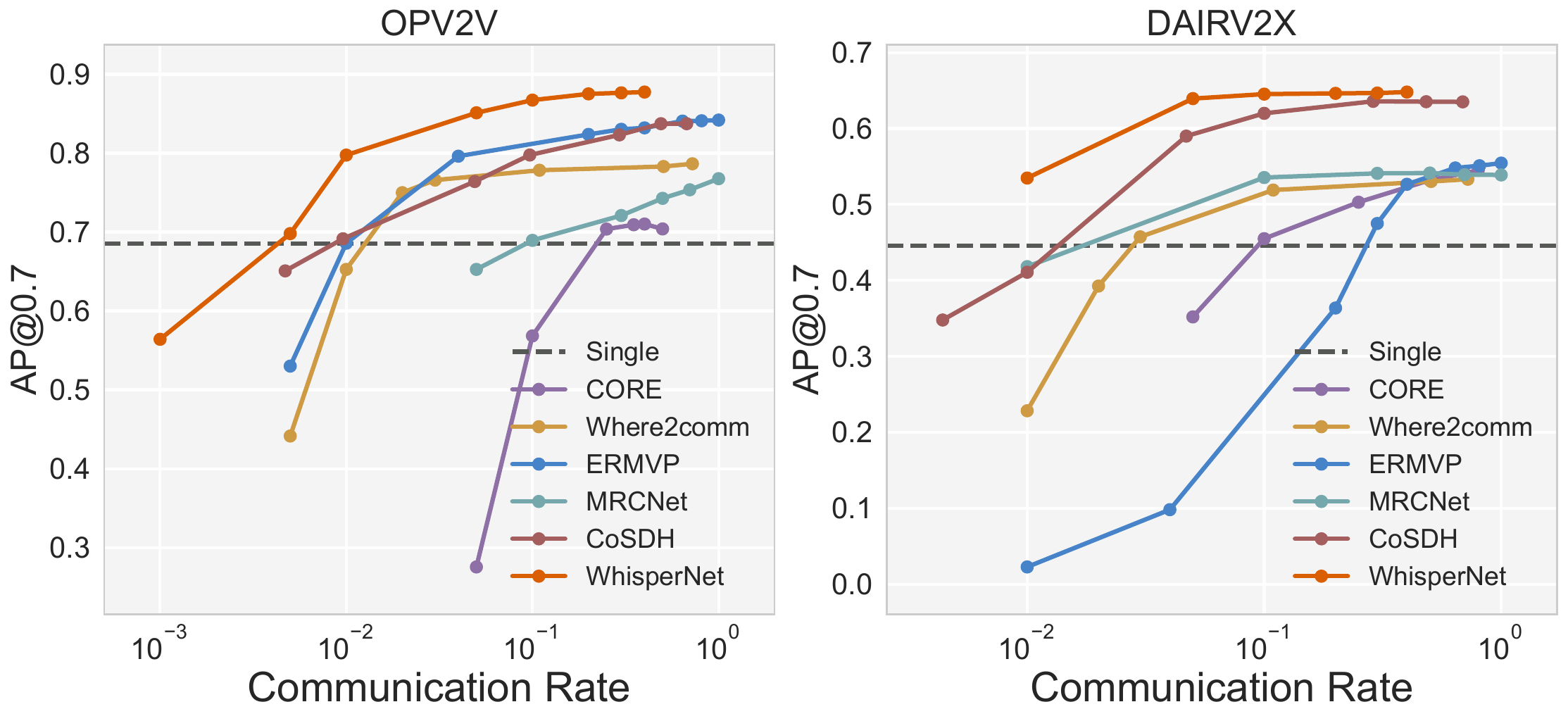} 
  \caption{Comparison of performance and bandwidth.}
  \label{fig:communication_comparison} 
\end{figure}

\subsection{Quantitative Evaluation}

\noindent \textbf{Benchmark Comparison.}
Tab.~\ref{tab:benchmark_comparison} presents the 3D object detection results on the OPV2V and DAIR-V2X benchmarks. Our proposed WhisperNet consistently achieves competitive or even state-of-the-art performance with a substantially smaller model size. Notably, on OPV2V, WhisperNet attains an AP@0.7 of 87.64\% while using only 7.28M parameters, which is significantly more compact than previous leading methods such as DSRC (10.02M). A similar trend is observed on DAIR-V2X, where WhisperNet outperforms the runner-up by 1.2\% AP@0.7 with fewer parameters. These results indicate that, despite being lightweight and designed to reduce redundant information exchange, WhisperNet can still effectively prioritize the most salient features by jointly modeling spatial and channel importance via its ego-centric coordination scheme.

\noindent \textbf{Performance-Bandwidth Trade-off.}
Fig.~\ref{fig:communication_comparison} shows that WhisperNet consistently achieves a leading trade-off between perception accuracy and communication cost. 
On OPV2V, WhisperNet outperforms the second-best competitor by 4.0\% at a 40\% communication rate. 
When the bandwidth is further reduced to only 1\%, this advantage becomes more pronounced, with performance gains of 10.6\% and 11.7\% on OPV2V and DAIR-V2X, respectively.
The bandwidth reduction capability of WhisperNet is further demonstrated by its ability to maintain baselines even when operating at only 0.5\% of the original bandwidth.  
In contrast, methods such as CORE, which rely on static feature compression, fail to adapt to extremely limited budgets.  
This advantage arises from WhisperNet's cooperative coordination mechanism, which enables all agents to determine what and where to transmit for optimal bandwidth utilization.

\begin{figure*}[ht] 
  \centering 
  \includegraphics[width=1.0\linewidth]{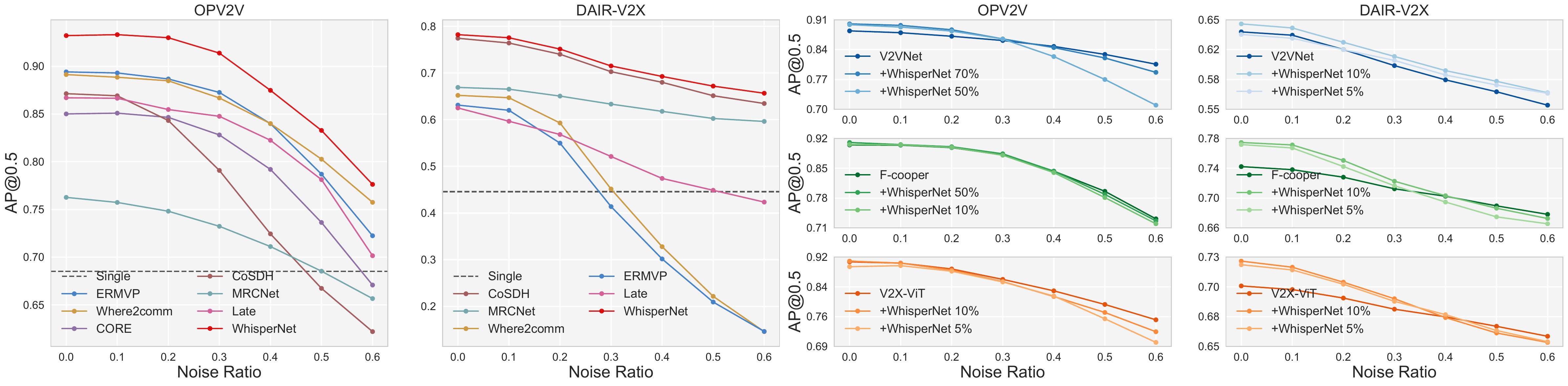}
  \caption{Comparison of model robustness to mixed noise. WhisperNet shows strong resilience compared to baselines, maintaining accuracy under high pose and heading perturbations}
  \label{fig:noise} 
\end{figure*}

\noindent \textbf{Plug-and-Play Evaluation.}
Fig.~\ref{fig:performance} validates WhisperNet's plug-and-play capability. When integrated into existing methods, it still offers a superior trade-off between accuracy and communication.
For 3D object detection, integrating WhisperNet allows these methods to maintain their full-bandwidth performance while using a fraction of the data. For instance, on OPV2V, V2X-ViT surpasses its baseline with only 10\% of the communication volume, while on DAIR-V2X, F-Cooper boosts its average AP from 0.704 to 0.723 with just 5\% of the original bandwidth. 
The advantage is particularly pronounced for dense prediction tasks like BEV semantic segmentation. At a 50\% communication rate, WhisperNet's average IoU is over 50\% higher than that of the purely spatial method Where2comm. This result highlights how our joint spatial-channel selection preserves the vital features that spatial methods discard, enabling superior performance at a minimal communication cost.

\noindent \textbf{Robustness to Noise.}
Fig.~\ref{fig:noise} evaluates methods' robustness to localization errors. We inject Gaussian noise into collaborator poses, including positional noise and heading noise, both sampled from $\mathcal{N}(0, \sigma^2)$.
As shown in the first two subfigures of Fig.~\ref{fig:noise}, WhisperNet demonstrates exceptional resilience compared to baselines. For instance, under a high noise level of $\sigma=0.6$ on DAIR-V2X, its performance degrades by only 16.0\%, in stark contrast to the over 75\% drop of methods like ERMVP. 
The remaining subfigures in Fig.~\ref{fig:noise} further show that this robustness can be transferred to other existing frameworks through WhisperNet's plug-and-play capability. The integrated models perform well under moderate noise and maintain respectable performance even under heavy noise. For instance, when integrated into V2X-ViT with just 5\% bandwidth, the enhanced model suffers a performance drop of less than 1\% under extreme noise, nearly matching the full-bandwidth baseline. These results confirm WhisperNet is both robust and capable of transferring this resilience to other frameworks while cutting bandwidth.

\subsection{Ablation Studies}


\noindent \textbf{Impact of Core Components.}  
Taking a transmission limit of under 50\% as an example, Tab.~\ref{tab:core_components} analyzes the contribution of major components in WhisperNet.
The Channel Merit Distributor (CMD) serves as the core mechanism for communication efficiency. It reduces bandwidth by half while maintaining 0.8532 in AP@0.7 on OPV2V. 
Further integrating the Spatial Focus Engine (SFE) enhances scene-level coordination, leading to consistent improvements across both datasets. 
Building upon this, the Collaborative Feature Routing (CFR) module performs channel-wise routing and matching. 
Finally, the full WhisperNet, which surpasses the baseline over 4.31\%/5.73\% on OPV2V/DAIR-V2X while cutting bandwidth by more than 50\%, demonstrates the synergy between our channel-spatial selection and collaborative routing design.

\begin{figure*}[ht] 
  \centering 
  \includegraphics[width=1.0\linewidth]{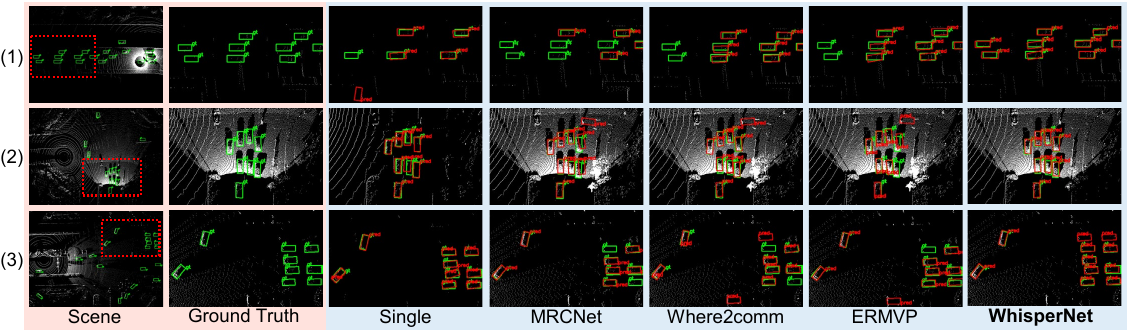}
  \caption{Visualization of mainstream methods in challenging driving scenarios.}
  \label{fig:vis_whispernet} 
\end{figure*}

  

\begin{table}[t]
  \centering
  \caption{Ablation on the contribution of core modules.}
  \label{tab:core_components}
  
  \begin{adjustbox}{max width=\linewidth}
    \begin{tabular}{ccccccc}
      \hline
      \multicolumn{3}{c}{Configuration} & \multicolumn{2}{c}{OPV2V} & \multicolumn{2}{c}{DAIR-V2X} \\
      \cmidrule(lr){1-3} \cmidrule(lr){4-5} \cmidrule(lr){6-7}
      CMD & SFE & CFR & Comm. & AP@0.5/0.7 & Comm. & AP@0.5/0.7 \\
      \hline
      & & & 32.81 & 0.8926/0.8333                             & 24.00 & 0.7532/0.5907 \\
      & & \ding{51} & 32.81 & 0.9284/0.8672                   & 24.00 & 0.7769/0.6315 \\
      \hline
      \ding{51} & & & 16.40 & 0.9206/0.8532                     & 12.00 & 0.7627/0.6036 \\
      \ding{51} & \ding{51} & & 15.62 & 0.9310/0.8683           & 11.27 & 0.7741/0.6271 \\
      \ding{51} & \ding{51} & \ding{51} & 15.62 & 0.9334/0.8764 & 11.27 & 0.7915/0.6480 \\
      \hline
    \end{tabular}
  \end{adjustbox}
\end{table}

\begin{table}[t]
\centering
\caption{Ablation on the size of the feature block.}
\label{tab:patch_selection}
\begin{adjustbox}{max width=\linewidth}
\begin{tabular}{cccc}
\hline
\multicolumn{4}{l}{AP@0.7 (OPV2V / DAIR-V2X)} \\
\hline
Patch Size & 5\% & 50\% & FPS \\
\hline
4 & 0.7732 / 0.5734 & 0.8752 / 0.6370 & 8.89 / 10.94 \\
2 & 0.8329 / 0.6305 & 0.8773 / 0.6415 & 8.19 / 10.14 \\
1 & 0.8511 / 0.6394 & 0.8764 / 0.6480 & 8.04 / 10.02 \\
\hline
\end{tabular}
\end{adjustbox}
\end{table}

\noindent \textbf{Impact of Patch Selection}
Tab.~\ref{tab:patch_selection} analyzes the impact of patch size on the accuracy-speed trade-off. At high bandwidth, a larger 4x4 patch offers a more efficient configuration, providing a 10\% FPS increase on OPV2V with only a negligible performance drop. However, this trend reverses under stringent communication constraints. At just a 5\% communication rate, the fine-grained 1x1 patch outperforms the 4x4 patch, boosting the AP@0.7 by over 7.8\%. This result demonstrates that a smaller patch size is critical for achieving fine-grained feature selection and preserving accuracy in low-bandwidth scenarios.

\begin{table}[t]
  \centering
  \caption{Ablation on the number of experts in our Collaborative Feature Routing module on OPV2V.}
  \begin{tabular}{cccc}
    \hline
    Experts & AP@0.5/0.7 & Params. (M) & GFLOPs \\
    \hline
    1 & 0.9158 / 0.8512 & 13.13 & 176.25 \\
    2 & 0.9275 / 0.8681 & 9.43  & 140.54 \\
    4 & 0.9334 / 0.8764 & 7.28  & 122.68 \\
    8 & 0.9121 / 0.8435 & 6.66  & 113.76 \\
    \hline
  \end{tabular}%
  \label{tab:expert_ablation}%
\end{table}%

\noindent \textbf{Analysis on the Number of Experts.} 
Tab.~\ref{tab:expert_ablation} reveals a clear trade-off in the number of experts used. Performance peaks at four experts, improving upon the single-expert by allowing feature channels to be partitioned for more specialized fusion. However, this trend reverses with eight experts, where performance collapses to 84.35\% due to over-specialization, which fragments information and hinders holistic context. We thus adopt four experts as the optimal configuration, as it strikes the best balance between performance, parameter count, and computational cost.

\noindent \textbf{Analysis of Channel Importance Metric.} 
Tab.~\ref{tab:channel_metric_ablation} ablates several channel scoring metrics to validate our use of the Laplacian magnitude. 
We find that metrics capturing high-frequency details through Laplacian responses consistently outperform those based on Jacobian or raw feature statistics (Max/Mean). 
For aggregation across channels, we evaluate different strategies and adopt a group-weighted formulation, where each channel is weighted by its semantic group importance before taking the maximum. 
This design achieves the highest 0.8729 AP@0.7, confirming that emphasizing high-frequency and high-confidence channels provides a more reliable indicator of perceptual importance for communication selection.

\begin{table}[t]
\centering
\caption{Ablation on channel importance metrics and patch score aggregation methods on OPV2V.}
\label{tab:channel_metric_ablation}
\begin{adjustbox}{max width=\linewidth}
\begin{tabular}{c | ccc | ccc}
\hline
\multicolumn{1}{c}{} & \multicolumn{3}{c}{Multi-Scale} & \multicolumn{3}{c}{Single-Scale} \\
\hline
Type & Weighted & Max & Mean & Weighted & Max & Mean \\
\hline
Laplacian & 0.8729 & 0.8709 & 0.8723 & 0.8402 & 0.8388 & 0.8401 \\
Jacobian & 0.8481 & 0.8467 & 0.8365 & 0.8345 & 0.8327 & 0.8337 \\
Abs & 0.8476 & 0.8454 & 0.8394 & 0.8317 & 0.8334 & 0.8317 \\
Max & 0.8705 & 0.8674 & 0.8669 & 0.8402 & 0.8317 & 0.8301 \\
LSD & 0.8533 & 0.8540 & 0.8481 & 0.8392 & 0.8316 & 0.8395 \\
Mean & 0.8474 & 0.8629 & 0.8623 & 0.8302 & 0.8017 & 0.8301 \\
\hline
\end{tabular}
\end{adjustbox}
\end{table}

\begin{figure}[t] 
  \centering 
  \includegraphics[width=1.0\linewidth]{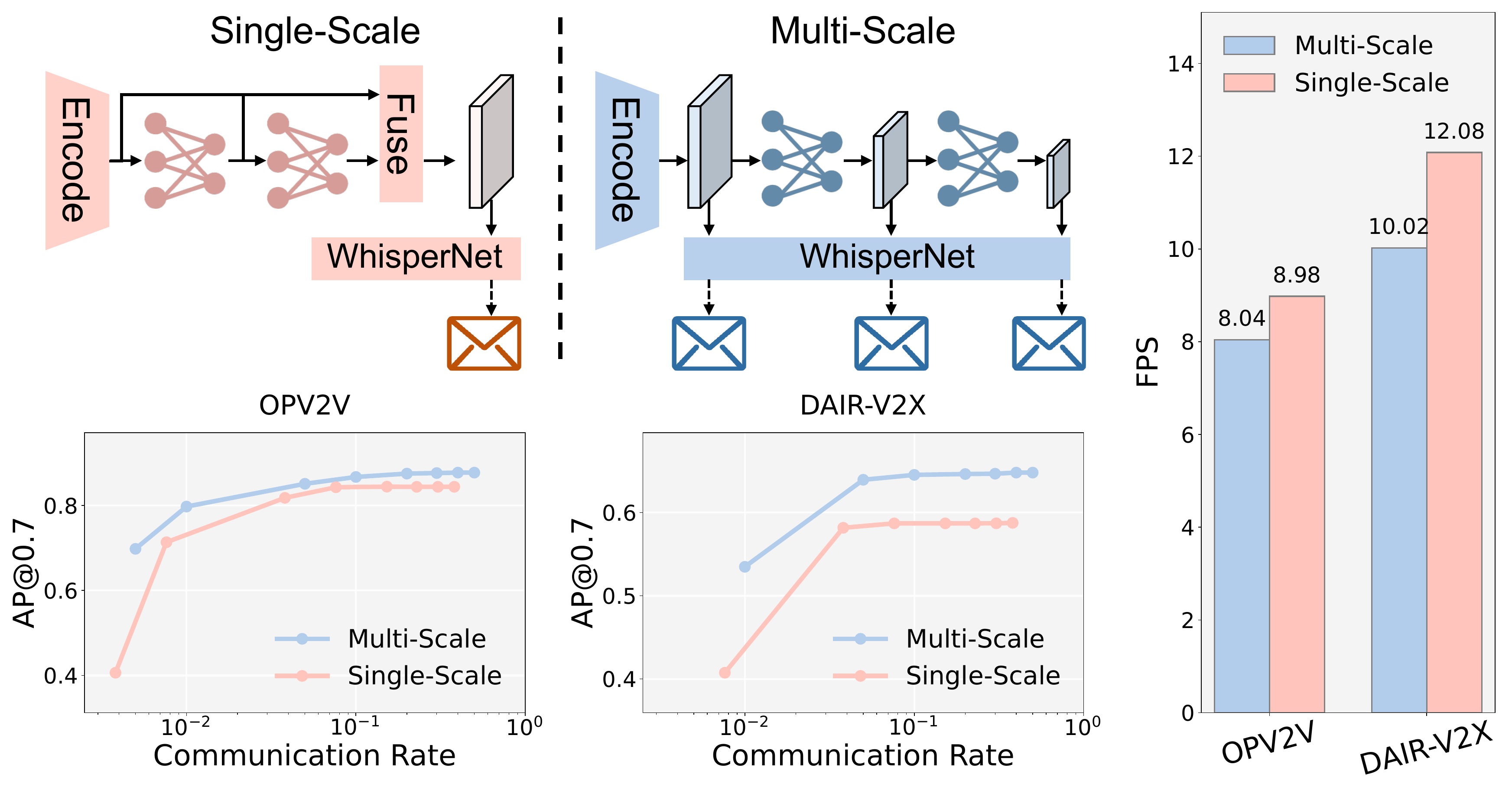}
  \caption{Ablation on single-scale and multi-scale feature transmission strategies.} 
  \label{fig:multisingle} 
\end{figure}

\noindent \textbf{Single vs. Multi-Scale Communication.}
Fig.~\ref{fig:multisingle} further ablates two feature transmission strategies within our framework: Single-Scale, which transmits a single fused intermediate feature, and Multi-Scale, which transmits a richer set of multi-scale features.
Multi-Scale strategy is more robust, while the Single-Scale approach is more efficient. The robustness advantage is most pronounced at a 1\%, where the Multi-Scale's AP@0.7 is over 31\% higher. Conversely, the Single-Scale approach reduces the data payload by 24\% and improves FPS by up to 21\%. Given the substantial robustness gains, we adopt the Multi-Scale strategy as WhisperNet's default configuration.

\subsection{Qualitative Visualization}


Fig.~\ref{fig:vis_whispernet} compares our method with baselines in three challenging scenarios featuring occlusion and long-range targets. The single-agent baseline predictably fails to perceive occluded objects. Other collaborative methods improve upon this but still exhibit frequent localization errors and missed detections. In all scenarios, WhisperNet generates significantly cleaner and more accurate results.

\section{Conclusion}

We presented WhisperNet, a bandwidth-aware collaborative perception framework that jointly optimizes spatial and channel dimensions through a two-sided communication strategy. 
Experiments show it achieves state-of-the-art performance on major benchmarks while preserving robustness to localization noise and maintaining high efficiency under severe bandwidth constraints. 
Its effectiveness arises from the coordinated design of sender-side importance estimation, receiver-side adaptive allocation, and collaborative feature alignment, offering a practical solution for scalable and communication-efficient multi-agent perception.

\noindent \textbf{Acknowledgement.} The work is supported in part by the S\&T Program of Hebei Province (Beijing-Tianjin-Hebei Collaborative Innovation Special Program) under Grant 25240701D.
   




{
    \small
    \bibliographystyle{ieeenat_fullname}
    \bibliography{main}
}

\end{document}